\documentclass[10pt,twocolumn,letterpaper]{article}
\pdfoutput=1
\usepackage{cvpr}
\usepackage{times}
\usepackage{epsfig}
\usepackage{graphicx}
\usepackage{subfigure}
\usepackage[export]{adjustbox}
\usepackage{amsmath}
\usepackage{amssymb}
\usepackage{tabularx} 
\newcolumntype{Y}{>{\centering\arraybackslash}X}

\usepackage{array,booktabs,arydshln,xcolor} 
\usepackage{amsthm}
\usepackage{algorithm}
\usepackage{algorithmic}
\usepackage{amsthm}
\usepackage{multirow}
\newtheorem{prop}{Proposition}

\cvprfinalcopy 

\def\cvprPaperID{2449} 

\ifcvprfinal\fi
\begin{document}

\title{SemiContour: A Semi-supervised Learning Approach for Contour Detection}
\author{Zizhao Zhang, ~Fuyong Xing, ~Xiaoshuang Shi, ~Lin Yang \\
	University of Florida, Gainesville, FL 32611, USA\\
	{\tt\small zizhao@cise.ufl.edu, \{f.xing,xsshi2015\}@ufl.edu, lin.yang@bme.ufl.edu}
	}


\maketitle

\begin{abstract}

Supervised contour detection methods usually require many labeled training images to obtain satisfactory performance. However, a large set of annotated data might be unavailable or extremely labor intensive. In this paper, we investigate the usage of semi-supervised learning (SSL) to obtain competitive detection accuracy with very limited training data (three labeled images). Specifically, we propose a semi-supervised structured ensemble learning approach for contour detection built on structured random forests (SRF). To allow SRF to be applicable to unlabeled data, we present an effective sparse representation approach to capture inherent structure in image patches by finding a compact and discriminative low-dimensional subspace representation in an unsupervised manner, enabling the incorporation of abundant unlabeled patches with their estimated structured labels to help SRF perform better node splitting. We re-examine the role of sparsity and propose a novel and fast sparse coding algorithm to boost the overall learning efficiency. To the best of our knowledge, this is the first attempt to apply SSL for contour detection. Extensive experiments on the BSDS500 segmentation dataset and the NYU Depth dataset demonstrate the superiority of the proposed method.

\end{abstract}

\vspace{-0.5cm}
\section{Introduction}
\label{introduction}
\begin{figure}[t]
	\begin{center}
		\includegraphics[width=0.49\textwidth]{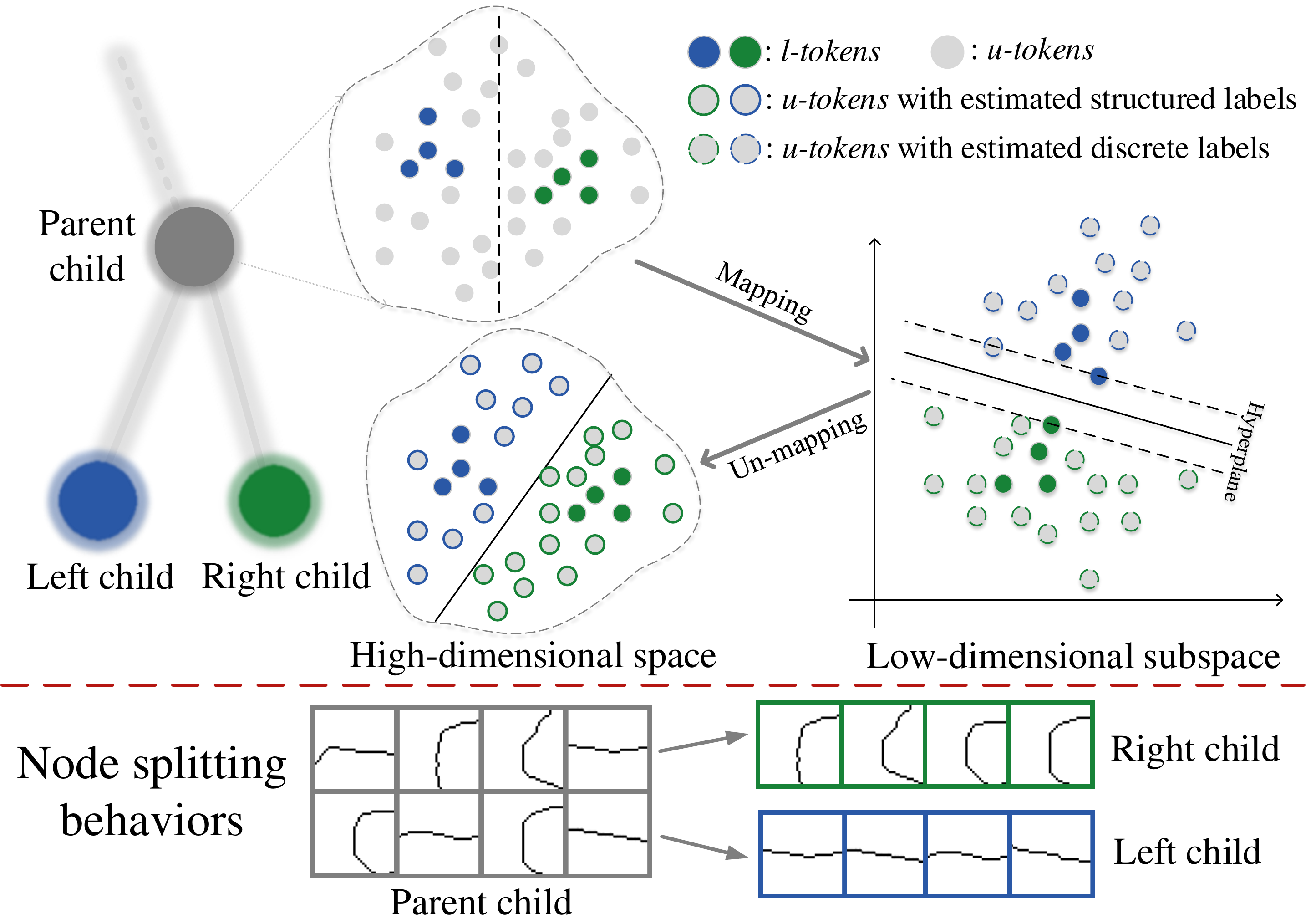}
		
	\end{center}
	\caption{Illustration of the proposed method. Top: At the parent node that contains \textit{u-tokens} and \textit{l-tokens} with corresponding structured labels, \textit{u-tokens} will help \textit{l-tokens} better estimate the separating plane for the node splitting. This is achieved by mapping tokens into a discriminative low-dimensional subspace and estimating \textit{u-tokens'} discrete labels. Then the \textit{u-tokens} are un-mapped to the original high-dimensional space associated with the estimated structured labels. Finally, all tokens will be propagated to child nodes. Bottom: A general view of the node splitting behavior to present how node splitting of SRF enables data with structured labels in the parent node to be categorized in child nodes.\vspace{-0.4cm}} \label{fig:outline}
\end{figure}
\par Contour detection is a fundamental but challenging computer vision task. In recent years, although the research of contour detection is gradually shifted from unsupervised learning to supervised learning, unsupervised contour detection approaches are still attractive, since it can be easily adopted into other image domains without the demand of a large amount of labeled data. However, one of the significant limitations is the high computational cost \cite{arbelaez2011contour,xiaofeng2012discriminatively}. On the other hand, the cutting-edge supervised contour detection methods, such as deep learning, rely on a huge amount of fully labeled training data, which often requires huge human efforts and domain expertise. Semi-supervised learning (SSL) \cite{tang2007co,liu2013semi,leistner2009semi} is an alternative technique to balance the trade-off between unsupervised learning and supervised learning. However, currently there exist no reports on semi-supervised learning based contour detection.

\par Supervised contour detection is often based on patch-to-patch or patch-to-pixel classification. Contours in local patches (denoted by sketch tokens \cite{lim2013sketch}) contain rich and well-known patterns, including straight lines, parallel lines, curves, T-junctions, Y-junctions, etc \cite{ren2006figure,lim2013sketch}. One of the main objectives of the most recent supervised contour detection methods is to classify these patterns using structure learning \cite{dollar2015pami,dollar2013structured}, sparse representation \cite{maire2014reconstructive, xiaofeng2012discriminatively}, convolution neutral network (CNN) \cite{shen2015deepcontour, ganin2014n,xie2015holistically}, etc. In our method, we use unsupervised techniques to capture the patterns of unlabeled image patches, enabling the successful training of the contour detector with a limited number of labeled images. For notation convenience, we denote labeled patches as \textit{l-tokens} and unlabeled patches as \textit{u-tokens}.

The proposed semi-supervised structured ensemble learning approach is built on structured random forests (SRF) \cite{kontschieder2011structured}. Inheriting from standard random forests (RF), SRF is popular because its: 1) fast prediction ability for high-dimensional data, 2) robustness to label noise \cite{liu2013semi}, and 3) good support to arbitrary size of outputs. However, similar to RF, SRF heavily relies on the number of labeled data \cite{leistner2009semi}. These properties make SRF a good candidate for SSL. 

In this paper, we propose to train SRF in a novel semi-supervised manner, which only requires a few number of labeled training images. By analyzing the learning behaviors of SRF, we observe that improving the node splitting performance for data with structured labels is the key for the successful training. To this end, we incorporate abundant \textit{u-tokens} into a limited number of \textit{l-tokens} to guide the node splitting, which is achieved by finding a discriminative low-dimensional subspace embedding using sparse representation techniques to learn a basis dictionary of the subspace in an unsupervised manner.  

In order to solve the sparse coding problem efficiently, we also propose a novel and fast algorithm to boost the overall learning efficiency. In addition, we demonstrate the max-margin properties of SRF, enabling us to use max-margin learning to dynamically estimate the structured labels for \textit{u-tokens} inside tree nodes. For better illustration, we explain the idea in Figure \ref{fig:outline}. In the experimental section, we show the vulnerability of other supervised methods to a limited number of labeled images and demonstrate that, with only 3 labeled images, our newly developed contour detector even matches or outperforms these methods which are fully trained over hundreds of labeled images.

\section{Related Works}
\label{relatedwork}
Recently, most advanced contour detection methods are based on strong supervision. Ren \etal use sparse code gradients (SCG) \cite{xiaofeng2012discriminatively} to estimate the local gradient contrast for gPb, which slightly improves the performance of gPb. Maire \etal \cite{maire2014reconstructive} propose to learn a reconstructive sparse transfer dictionary to address contour representation. These methods indicate the strong capability of sparse representation techniques to capture the contour structure in image patches. In the ensemble learning family, Lim \etal \cite{lim2013sketch} propose sketch tokens, a mid-level feature representation, to capture local contour structure, and train a RF classifier to discriminate the patterns of sketch tokens. Doll{\'a}r \etal \cite{dollar2013structured} propose a structured edge (SE) detector that outperforms sketch tokens by training a SRF classifier instead. Several variants of SRF are also successfully applied to image patch classification \cite{shen2015deepcontour,dollar2015pami,myers2015affordance,arbelaez2014multiscale,fujun2015maccai,teo2015fast}. Recently, CNN has shown its strengths in contour detection \cite{shen2015deepcontour, ganin2014n,bertasius2014deepedge}, and its success is attributed to the complex and deep networks with new losses to capture contour structure. One major drawback of CNN, as well as other supervised learning methods, is its high demand of labeled data.

Semi-supervised learning (SSL) has been studied to alleviate the aforementioned problems \cite{chapelle2006semi,leistner2009semi,liu2013semi,tang2007co}. Leistner \etal \cite{leistner2009semi} treat unlabeled data as additional variables to be jointly optimized with RF iteratively. Liu \etal \cite{liu2013semi} instead use unlabeled data to help the node splitting of RF and obtain improved performance. However, it is difficult for these methods to avoid the curse of dimensionality. By contrast, this paper takes advantage of several properties of SRF to achieve an accurate contour detector with very few labeled training images. We address several critical problems to successfully learn SRF in a semi-supervised manner without much sacrificing the training and testing efficiency by 1) estimating the structured labels for \textit{u-tokens} lying on a complex and high-dimensional space, and 2) preventing noises of extensively incorporated \textit{u-tokens} from misleading the entire learning process of SRF. 

\vspace{-0.2cm}
\section{SSL Overview in Contour Detection}
\par SSL uses a large number of unlabeled data $\mathcal{D}^U=\{x \in \mathcal{X}\}$ to augment a small number of labeled data $\mathcal{D}^L=\{(x, y) \in \mathcal{X} \times \mathcal{Y} \}$ and learns a prediction mapping function $f: \mathcal{X} \mapsto \mathcal{Y}$. In the scenario of contour detection, we denote $x$ as a token, and $y$ as its corresponding structured label of a certain pattern.

\par Contour detection performance of supervised methods is not only determined by the number of \textit{l-tokens} in $\mathcal{D}^L$, but also affected by the number of labeled images, from which \textit{l-tokens} are sampled \cite{dollar2015pami}. This is because the limited information in \textit{l-tokens} sampled from a few labeled images is severely biased, which can not lead to a general classification model. On the contrary, sufficient \textit{u-tokens} in $\mathcal{D}^U$ sampled from many unlabeled images contain abundant information that is easy to acquire. We apply SSL to take advantage of \textit{u-tokens} to improve the supervised training of our contour detector. However, \textit{u-tokens} always have large appearance variations, so it is difficult to estimate their structured labels in the high-dimensional space $\mathcal{Y}$. 

\begin{figure}[t]
	\begin{center}
		
		\subfigure[Mean patterns of 200 images]{\includegraphics[width=0.48\linewidth]{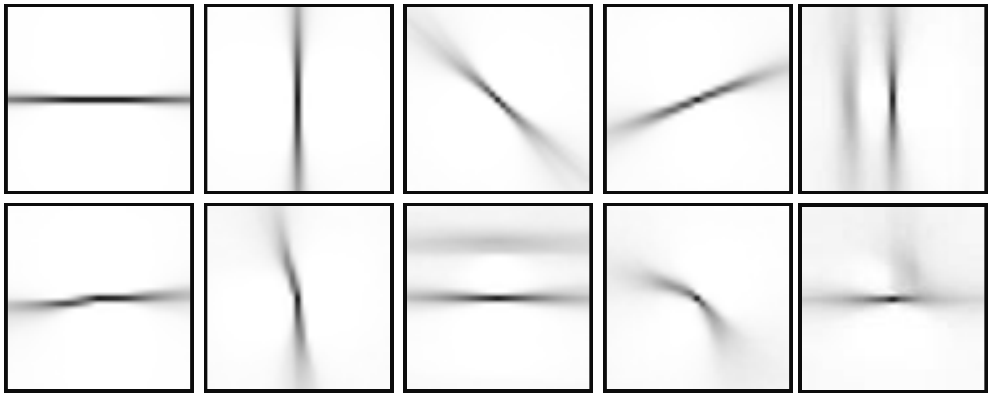}}
		\subfigure[Mean patterns of 3 images]{\includegraphics[width=0.48\linewidth]{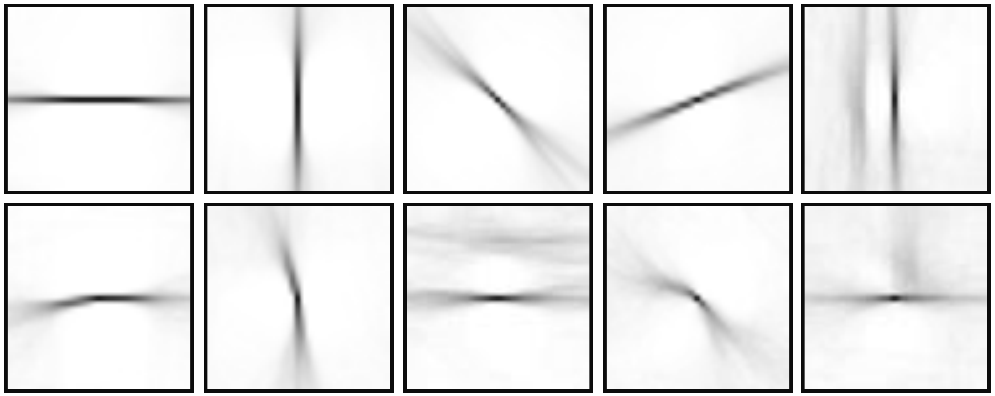}}
	\end{center}
	\vspace{-0.4cm}
	\caption{Examples of mean patterns calculated by clustering the structured labels of tokens sampled from 200 images 
		and 3 images into 150 classes \cite{lim2013sketch}. The patterns calculated from 3 images are almost identical to the patterns calculated from 200 images.\vspace{-0.3cm}} \label{fig:dist}
\end{figure}

We propose to estimate the structure labels of \textit{u-tokens} by transferring existing structured labels of \textit{l-tokens}. Because the patterns of the structured labels are limited and shared from images to images, which can be categorized into a finite number of classes (e.g., straight lines, parallel lines, and T-junctions), the structured labels of \textit{l-tokens} from a few images are sufficient to approximate the structured labels of massive \textit{u-tokens} from many images. We demonstrate this in Figure \ref{fig:dist}.

\section{SSL via Structured Ensemble Learning}
\label{structureleanring}
In this section we describe the proposed semi-supervised ensemble learning approach for contour detection. The method is built on the structured random forests (SRF), which has a similar learning procedure as the standard random forest (RF) \cite{breiman2001random}. The major challenge of training SRF is that structured labels usually lie on a complex and high-dimensional space, therefore direct learning criteria for node splitting in RF is not well defined. Existing solutions \cite{kontschieder2011structured,dollar2015pami} can only handle fully labeled data, and are not applicable in our case that contains both unlabeled and labeled data. We will start by briefly introducing SRF and analyze several favorable properties of SRF for SSL, and then present the proposed SSL based contour detection method.
\subsection{Structured random forest}
\label{sec:srf}
SRF is an ensemble learning technique with structured outputs, which ensembles $T$ independently trained decision trees as a forest $\mathcal{F} = \{F_t\}_{t=1}^T$. Robust SRF always has large diversity among trees, which is achieved by bootstrapping training data and features to prevent overfitting. Given a set of training data $\mathcal{D}$, starting from the root node, a decision tree $F_t$ attempts to propagate the data from top to bottom until data with different labels are categorized in leaf nodes.

Specifically, for all data $x \in \mathcal{D}_i$ in node $i$, a local weak learner $h(x, \theta)=\textbf{1}[x_k < \tau ]$ propagates $x$ to its left substree if $h(\cdot)=1$, and right substree otherwise. $\theta=(\tau, k)$ is learned by maximizing the information gain $\mathcal{I}_i$:
\vspace{-0.1cm}
\begin{equation} \vspace{-0.1cm}
\theta^\star = \operatorname*{argmax}_{\tau \in \mathbb{R}, k \in \mathbb{Z}} \; \mathcal{I}_i.\\
\end{equation}
The optimization is driven by the Gini impurity or Entropy \cite{breiman2001random}.
$y \in \mathcal{Y} = \mathbb{Z}^{m\cdot m}$ is a structured label with the same size as the training tokens. To enable the optimization of $\mathcal{I}_i$ for structured labels, Doll{\'a}r \etal \cite{dollar2013structured} propose a mapping $\Pi: \mathcal{Y} \mapsto \mathcal{L}$ to project structured labels into a discrete space, $l\in \mathcal{L} = \{1,...,Z\}$, and then follow the standard way. The training terminates (i.e., leaf nodes are reached) until a stopping criteria is satisfied \cite{breiman2001random}. The most representative $y$ (i.e., closet to mean) is stored in the leaf node as its structured prediction, i.e., the posterior $p(y|x)$.

The overall prediction function of SRF ensembles $T$ predictions from all decision trees, which is defined as	
\vspace{-0.1cm}
\begin{equation}\vspace{-0.1cm}
\operatorname*{argmax}_{y \in \mathcal{Y}} p(y|x,\mathcal{F}) = \frac{1}{T} \sum_{t=1}^{T} \operatorname*{argmax}_{y \in \mathcal{Y}} p(y|x,F_t).
\end{equation}
To obtain optimal performance, given a test image, we densely sample tokens in multi-scales so that a single pixel can get $m \times m \times T\times \text{(\# of scales)}$ predictions in total. The structured outputs force the spatial continuity. The averaged prediction yields soft contour responses, which intrinsically alleviate noise effects and indicate a good sign to performing SSL in SRF. 

Good features play an important role in the success of SRF. Shen \etal \cite{shen2015deepcontour} improve the SE contour detector \cite{dollar2013structured} by replacing the widely used HoG-like features with CNN features. In fact, this CNN classifier itself is a weak contour detector used to generate better gradient features.  
Inspired by this idea, we use a limited number of \textit{l-tokens} from a few labeled images to first train a weak SE contour detector (denoted by $\Gamma$) \cite{dollar2013structured}. $\Gamma$ produces efficient detection and provides prior knowledge for \textit{u-tokens} to facilitate SSL. We will see its further usage subsequently. In our method, we use three color channels ($Luv$), two gradient magnitude (obtained from $\Gamma$) and eight orientation channels in two scales, and thus the total feature space is $\mathcal{X} \in \mathbb{R}^{m\cdot m\cdot 13}$, which is similar to the configuration in \cite{lim2013sketch}.

\subsection{Semi-supervised SRF learning}
\label{ssl_srf}
In our method, maximizing the information gain $\mathcal{I}_i$ is achieved by minimizing the Gini impurity measurement $G$ \cite{criminisi2012decision}, which is defined as	
\vspace{-0.1cm}
\begin{equation} \vspace{-0.1cm}
	G(\mathcal{\tilde{D}}_{i}) =  \sum_{j=1}^{Z} p_j(l|x_k)(1-p_j(l|x_k)),
\end{equation}
where $p_j(l|x_k)$ denotes the label empirical distribution of class $j$ in $\mathcal{\tilde D}_{i}$ with respect to the $k$-th feature dimension. We adopt the mapping function $\Pi$ \cite{dollar2013structured} to map structured labels of \textit{l-tokens} to discrete labels. $\mathcal{\tilde D}_{i}=\{(x,l)|(x,y)\in \mathcal{D}_{i}, l=\Pi(y)\}$ denotes $\mathcal{D}_i$ when $x$ is with the discrete label. Intuitively, minimizing $G$ is to find a separating line in the $k$-th feature dimension (several feature dimensions can be used together to define a separating hyperplane \cite{criminisi2012decision}) to split $\mathcal{\tilde D}_i$ in the whole feature space into the left and right subtrees, so that $p_j$ on both sides are maximized \cite{breiman2001random}. Proposition 1 proves the close relationship of the Gini impurity to the max-margin learning.

\begin{prop} Given the hinge loss function $\xi$ of max-margin learning, the Gini impurity function $\sum_{\mathcal{L}} p_j(1-p_j)$ is its special case.
\begin{proof} \let\qed\relax 
Since $l(w^Tx) \geq 0$, if $l(w^Tx) \leq 1$, then we have:
\vspace{-0.25cm}
\begin{equation*} \vspace{-0.3cm}
\normalsize 
\begin{split}
\xi(\mathcal{\tilde D}_i) &= \frac{1}{|\mathcal{\tilde D}_i|}  \sum_{(x,l) \in \mathcal{\tilde D}_i} \sum_{j=1}^{Z} \textbf{1}[l=j] \max(0,1-l(w^Tx)) \\
& =  \sum_{j=1}^{Z} p_j (1-l(w^Tx)),
\end{split}
\end{equation*} 
where $p_j = \frac{\sum_{(x,l) \in \mathcal{\tilde D}_i} \textbf{1}[l=j]}{|\mathcal{\tilde D}_i|}$. Because $p_j \propto l(w^Tx)$, the Proposition holds. A generalized theorem is given in \cite{leistner2009semi}.
\end{proof}
\end{prop}

\noindent
\textbf{Incorporate Unlabeled Data} It is well-known that a limited number of labeled data always lead to biased max-margin estimation. We incorporate \textit{u-tokens} into the limited number of \textit{l-tokens} to improve the max-margin estimation of weak learners in every node. However, $p(l|x^u)$ of \textit{u-tokens} is unavailable for computing Gini impurity. One solution to address this problem \cite{liu2013semi} is to apply a kernel density estimator to obtain $p(x^u|l)$ and use the Bayes rule to obtain $p(l|x^u)$. In this approach, a proper selection of bandwidth is not trivial. In addition, it can not handle structure labels and the high-dimensional space, on which \textit{u-tokens} lie. In our method, we propose to map tokens into a more discriminate low-dimensional subspace associated with discrete labels using a learned mapping $\mathcal{S}$, and find a hyperplane $w$ to estimate $p(l|x^u)$. In this scenario, the goal is to calculate the bases of the subspace. The data correlation in the subspace is consistent with that in the original space so that the estimated $p(l|x^u)$ will not mislead the weak learners. In Section \ref{sec:str}, we demonstrate that this goal can be achieved using sparse representation techniques. 
\vspace{0.3em}

\noindent
\textbf{SRF Node Splitting Behaviors} During the training stage of SRF, tokens with various patterns are chaotic in the top level nodes, and weak learners produce coarse splitting results; while at the bottom level nodes, the splitting becomes more subtle. For example, suppose $l\in \{0,1\}$, the weak learner in the root node intends to split foreground and background tokens into the left and right subtrees, respectively. The top level nodes tend to split the straight line and broken line patterns, whereas weak learners tend to split $40$ degree and $30$ degree straight lines in the bottom level nodes, in which patterns are more pure. Considering this property, we propose a novel dynamic structured label transfer approach to estimate the structured labels for \textit{u-tokens}.
\begin{figure*}[t]
	\begin{center}
		\includegraphics[width=0.99\textwidth]{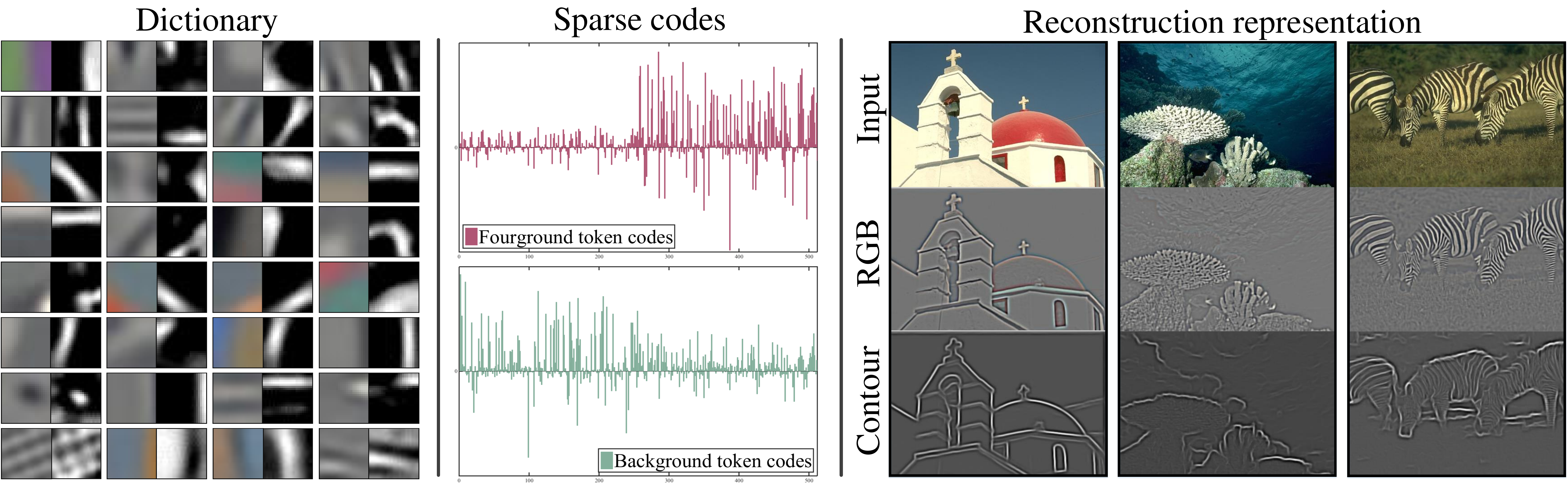}
		
	\end{center}
	\vspace{-0.5cm}
	\caption{Illustration of the sparse token representation. We empirically set $V{=}256$ and $K{=}3$ when training $M_{f}$ and $M_b$ separately, in which $1\times 10^5$ training tokens are used for each. The overall dictionary is $M = [M_{b}, M_{f}] \in \mathbb{R}^{(m\cdot m\cdot 4) \times 512}$ and the sparsity is set as $K{=}6$. Left: Examples of some bases with the RGB  channels (left) and the contour  channel (right). Middle: Average sparse code ($512$ dimensional vectors) values of foreground tokens (top) and background tokens (bottom) from test images. We can see foreground tokens tend to select bases (i.e., assigning high weights to bases) belonging to $M_{f}$,  while background tokens tend to select bases belonging to $M_{b}$. Right: Reconstruction representation of input images (top) with the RGB channels (middle) and the contour channel (bottom). Contours are well preserved and background noises are greatly suppressed.} \label{fig:sparse}
\end{figure*}

\subsection{Dynamic structured label transfer}
\label{dynamic-transfer}
Because it is challenging to directly estimate high-dimensional structured labels for \textit{u-tokens}, in our method, we transfer existing structured labels of \textit{l-tokens} to \textit{u-tokens}. An important concern is to prevent inaccurate structured label estimation for \textit{u-tokens} from destroying the learning of SRF. Suppose we have mapped tokens in node $i$ into a low-dimensional subspace using $\mathcal{S}$, we first search for a max-margin hyperplane $w$ using a linear wighted binary support vector machine trained over \textit{l-tokens} with discrete labels in this node (so the number of discrete labels $Z{=}2$ in our case). In this way, for an \textit{u-token} $x^u$, we can estimate its discrete label (i.e., $p(l|x^u)$) through $sigmoid\left(w^T\mathcal{S}(x^u)\right)$. 

To estimate its structured label, we adopt the nearest search to find the best match in the candidate pool of \textit{l-tokens} with the same discrete label as $x^u$. The structured label transfer function $\mathcal{H}: \mathcal{X} \mapsto \mathcal{Y}$ is defined as
\begin{equation}
\begin{split}
y^\star = \mathcal{H}(x^u) = \operatorname*{argmin}_{(x,y) \in \mathcal{D}_i^{L} } \textit{dist}(\mathcal{S}(x), \mathcal{S}(x^u)),
\end{split}
\end{equation}
where $\textit{dist}(\cdot,\cdot)$ is the cosine metric. In Section \ref{fastsparse}, we will see that $\mathcal{S}$ generates very sparse low-dimensional representation for tokens so that the steps of finding the hyperplane and performing the nearest search are computationally efficient. Finally we can easily map \textit{u-tokens} associated with structure labels back to their original space, and all tokens in node $i$ are propagated to child nodes. 

The brute-force searching at the top level nodes may yield inaccurate structured label estimation for \textit{u-tokens} due to the chaotic patterns and coarse discrete labels. In addition, it might lead to unnecessary computations because of redundant structured labels in one class. To tackle these problems, we dynamically update the transferred structured labels during the training of SRF. At the root node, we transfer initial structured labels to \textit{u-tokens}. As the tree goes deeper, weak learners gradually purify the candidate pool by decreasing the token volume and pattern variety. Therefore, the dynamically estimated structured labels of \textit{u-tokens} will become more reliable in the bottom level nodes. Since the number of \textit{u-tokens} is much larger than that of \textit{l-tokens}, some bottom level nodes might contain less or no \textit{l-tokens}. We treat \textit{u-tokens} with high probability as \textit{l-tokens} when a node does not contain enough \textit{l-tokens}, less than $10$ in our case. In addition, we randomly pick a subset instead of the entire candidate pool to perform the nearest search in each individual node.

\section{Sparse Token Representation}
\label{sec:str}
This section discusses the approach of finding the subspace mapping $\mathcal{S}$ mentioned in Section \ref{ssl_srf}. We first describe how to learn a token dictionary to construct the bases of the low-dimensional subspace, and then present a novel and fast sparse coding algorithm to accelerate the computation of $\mathcal{S}$.

\subsection{Sparse token dictionary}
\label{sec:dic}
Sparse representation has been proven to be effective to represent local image patches \cite{maire2014reconstructive,xiaofeng2012discriminatively,liu2011robust}. In our method, we pursue a compact set of the low-level structural primitives to describe contour patterns by learning a token dictionary. Specifically, any token $x$ can be represented by a linear combination of $K$ bases in a dictionary $M=[m_1,...,m_V]$ containing $V$ bases ($K \ll V$). A sparse code $c$ is calculated to select the $K$ bases. Given a set of training tokens $X=[x_1,...,x_n]$, the dictionary $M$, as well as the associated $C=[c_1,...,c_n]$, is learned by minimizing the reconstruction error \cite{aharon2006img}: 
\begin{algorithm}[t]
	\caption{Fast sparse coding}\label{solver}
	\begin{algorithmic}[1] 
		\renewcommand{\algorithmicrequire}{\textbf{Input:}}
		\renewcommand{\algorithmicensure}{\textbf{Output:}}
		\REQUIRE A target data $x \in \mathbb{R}^{d}$, a dictionary $M\in \mathbb{R}^{d\times V}$, and a sparsity value $K$
		\ENSURE A sparse code $c \in \mathbb{R}^{V}$ \\
		\FOR {$v = [1,2,...,V]$}
		\STATE Obtain the score $s_v$ for $m_v$:
		\vspace{-0.5em}
		\begin{equation}
		\begin{split}
		s_v^\star &= \operatorname*{argmin}_{s_v} \|x - m_v s_v\|_2^2 \\
		& = (m_v^Tm_v)^{-1}\sqrt{m_v^Tyy^Tm_v}
		\end{split}
		\end{equation}
		\vspace{-1.0em}
		\ENDFOR
		\STATE 
		 Construct a small size dictionary matrix $M_s \in \mathbb{R}^{d\times K}$ using the $K$ bases associated with the first $K$ largest scores in  $[s_1,...,s_V]$
		\STATE Solve a constrained leasts-squares problem:
		\vspace{-0.5em}
		\begin{equation}	
		\begin{split}
		c_s^\star & =\operatorname*{argmin}_{c_s} \|x - M_sc_s \|^2_2 + \lambda\|c_s\|_2^2 \\
		& =  (M_s^T M_s+\lambda I)^{-1} M_s^Tx, \;c_s \in \mathbb{R}^{K}.
		\end{split}
		\end{equation}
		\vspace{-0.5em}
		\STATE Obtain a sparse code $c$ by filling its $K$ entries with $c_s^\star$ indexed by $s$
		\RETURN{The sparse code $c$} 
	\end{algorithmic} 
\end{algorithm} 
\begin{equation}
\label{eqn:sc}
\begin{split}
&  \operatorname*{argmin}_{M,C}\| X - MC \|_{\small F}^2, \\
 s.t.\;\forall i, \;&\|m_i\|_2 = 1 \; and \;\,\forall j, \; ||c_j||_0 \leq K, \\
\end{split}
\end{equation}
where $\|\cdot \|_0$ is \(\ell_{0}\)-norm, $\|c_j\|_0 = \sum_i \mathbf{1} [c_{ji}\neq 0]$, to ensure that a sparse code $c$ only has $K$ nonzero entries. $\|\cdot \|_F$ is the Frobenius norm.  
Inspired by \cite{maire2014reconstructive}, we adopt MI-KSVD, a variant of the popular K-SVD, to solve Eqn. (\ref{eqn:sc}) for better sparse reconstruction \cite{bo2013multipath}. However, the dictionary $M$ is learned in an unsupervised manner, so it is not task-specific and its learning performance can be influenced by large appearance variances in tokens from different images. In particular, we observe that the cluttered background tokens (i.e., tokens contain no annotated contour inside) may cause unnecessary false positives. To ameliorate these problems, we introduce the prior label knowledge as an extra feature in the dictionary to improve its learning performance. 
\par Specifically, for an RGB token, we apply $\Gamma$ (Section \ref{sec:srf}) to generate its corresponding contours as the prior label knowledge, i.e., a patch with detected contours. In this way, the new featured token $x$ will have $4$ channels, which is represented as $x = [x^{(r)}, x^{(g)}, x^{(b)}, x^{(e)}]^T \in \mathbb{R}^{m\cdot m \cdot 4}$, where $x^{(e)}$ is the contour channel corresponding to the RGB channels ($x^{(r)}$, $x^{(g)}$, and $x^{(b)}$). We model background with $M_{b}$ and foreground with $M_{f}$, respectively. Figure \ref{fig:sparse} illustrates how the dictionary represents the structure in tokens.

\par In our method, both \textit{u-tokens} and \textit{l-tokens} are used as the training data for dictionary learning, which are sampled from unlabeled and labeled images, respectively. Foreground tokens are extracted if they straddle any contours indicated by the ground truth. The rest are background tokens.  Because the ground truth of \textit{u-tokens} is unavailable, we use the probability outputs of $\Gamma$ to help us sample high confident foreground and background \textit{u-tokens}. 

\begin{figure*}[t]
	\begin{center}
		\includegraphics[width=.9999\linewidth]{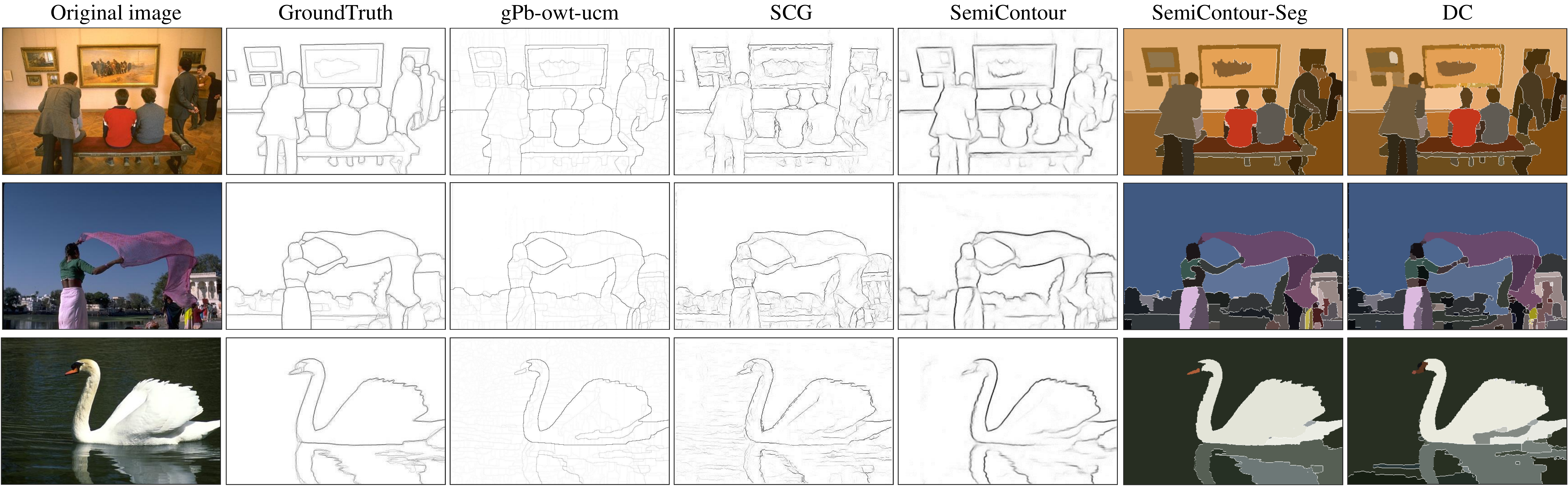}
	\end{center}
	\vspace{-0.3cm}
	\caption{Experimental results on BSDS500. The first two columns show the original image and ground truth. The next three columns show results of comparative methods and our SemiContour. SemiContour produces more clean background and stronger responses on high confident contours as indicated by ground truth. The last two columns show the segmentation results of SemiContour-Seg and DC \cite{donoser2014discrete}. Our method produces more consistent segmentation due to less false positive contour detection. } \vspace{-0.3cm}\label{fig:edge_vis}
\end{figure*}

\subsection{Subspace mapping using fast sparse coding}
\label{fastsparse}
As we mentioned in Section \ref{ssl_srf}, we use the mapping function $\mathcal{S}$ to provide a compact and discriminative low-dimensional representation for a token $x$. Given a learned dictionary $M$ in Section \ref{sec:dic}, the subspace representation of $x$ is defined as 
\begin{equation}
\label{eq:coding}
\begin{split}
\mathcal{S}(x) = c^\star = \operatorname*{argmin}_{c} \|x - Mc\|^2, \quad s.t.\, \|c\|_0 \leq K.
\end{split}
\end{equation}
It is well-known that solving Eqn. (\ref{eq:coding}) is NP-hard (\(\ell_{0}\)-norm). One typical algorithm to solve this problem is orthogonal matching pursuit (OMP) \cite{pati1993orthogonal}. Many other algorithms often relax it to the tractable \(\ell_{1}\)-norm minimization problem. Yang \etal \cite{yang2012beyond} show that \(\ell_{1}\)-norm provides better classification meaningful information than \(\ell_{0}\)-norm. The main reason is that, unlike \(\ell_{0}\)-norm that only selects the dictionary bases, \(\ell_{1}\)-norm also assigns weights to the selected bases to determine their contributions. Usually, high weights are often assigned to the bases similar to the target data \cite{wang2010locality}. In this paper, we propose a novel and fast sparse coding algorithm, which is scalable to a large number of target data. 

Based on the above observation, we approximate the computation of sparse coding by two steps: 1) basis selection, which measures the similarity score of each basis to the target data individually and then selects the bases with large scores; 2) reconstruction error minimization, which aims to assign weights to selected bases. The details are summarized in Algorithm \ref{solver}. Given a target data, we first compute a sequence of scores with respect to each basis (steps 1 to 3). Next we select $K$ bases associated with the first $K$ largest scores to construct a small size dictionary $M_s$ (step 4). Then we solve a constrained least-squares problem to obtain the coefficient $c_s$ and assign weights to the selected bases (step 5). The regularization parameter $\lambda$ is set to a small value, $10^{-4}$. Finally, the value of $c_s$ is mapped to $c$ as the final sparse code (steps 6 to 7). 

Unlike many existing methods, our proposed algorithm decouples the sparse coding optimization to problems with analytical solutions, which do not need any iteration. Therefore, our algorithm is faster than others that directly solve \(\ell_{0}\)-norm or \(\ell_{1}\)-norm problems.

\section{Experimental Results}
In this section, we first evaluate the performance of the proposed method for contour detection on two public datasets, and then compare the efficiency of the proposed sparse coding solver with several state-of-the-arts.

\newcommand{\scdot}{{\cdot}}
\subsection{Contour detection performance}
\label{sec:cdp}
We test the proposed approach on the Berkeley Segmentation Dataset and Benchmark (BSDS500) \cite{arbelaez2011contour} and the NYUD Depth (NYUD) V2 Dataset \cite{silberman2012indoor}. We measure the contour detection accuracy using several criteria: F-measures with fixed optimal threshold (ODS) and per-image threshold (OIS), precision/recall (PR) curves, and average precision (AP) \cite{arbelaez2011contour}. In all experiments, we use tokens with a size of $m{=}12$ based on an observation that a larger size (e.g., $m{=}30$) will significantly reduce the sparse representation performance, while a smaller size (e.g., $m{=}5$) can hardly represent rich patterns. This token size is also adopted by SRF to train $T{=}10$ trees. The skeleton operation is applied to the output contour images of the proposed SemiContour using the non-maximal suppression for quantitative evaluation.
\vspace{0.1em}

\noindent
\textbf{Training Image Settings:} We randomly split training images into a labeled set and an unlabeled set\footnote{To compensate for possible insufficient foreground \textit{l-tokens}, we duplicated images in the labeled set by histogram matching.}. We use a fair and relative large number of training tokens for all comparative methods, i.e., $1\times 10^5$ for background and foreground. Tokens (including \textit{l-tokens} and \textit{u-tokens}) are evenly sampled from each image in both sets. $\Gamma$ is trained over the labeled set to sample \textit{u-tokens} from the unlabeled set. Three tests are performed and average accuracies are reported as the final results.

\begin{table}[t]
	\caption{Contour detection results on BSDS500. In the first column, from top to down, the first block is the human annotations; the second block is unsupervised methods; the third block is supervised methods; the fourth block is our methods. For supervised methods (third block), we show the performance using both $3$ and $200$ training images (shown in $(\cdot)$).}\label{table:bsds}
	\vspace{-0.35cm}
	\begin{center}
		\begin{tabularx}{.47\textwidth}{c|YYY}
			                    ~                      & ODS           & OIS          & AP            \\ \specialrule{1.5pt}{0pt}{0pt}
			                  Human                    & .80           & .80          & -             \\ \hline
			Canny       \cite{canny1986computational}  & .60           & .64          & .58           \\
			Felz-Hutt \cite{felzenszwalb2004efficient} & .61           & .64          & .56           \\
			 Normalized Cuts \cite{cour2005spectral}   & .64           & .68          & .48           \\
			   Mean Shift  \cite{comaniciu2002mean}    & .64           & .68          & .56           \\

			   Gb  \cite{leordeanu2014generalized}     & .69           & .72          & .72           \\

			 gPb-owt-ucm  \cite{arbelaez2011contour}   & .73           & .76          & .70           \\ \hline
			        ISCRA \cite{ren2013image}          & - 	  (.72)    & -    (.75)   & -    (.46)    \\
			    Sketch Tokens \cite{lim2013sketch}     & .64(.73)      & .66(.75)     & .58(.78)      \\

			 SCG  \cite{xiaofeng2012discriminatively}  & .73(.74)      & .75(.76)     & .76(.77)      \\
			      SE \cite{dollar2013structured}       & .66(.74)      & .68(.76)     & .69(.78)      \\
			       SE-Var \cite{dollar2015pami}        & .69(.75)      & .72(.77)     & .74(.80)      \\ \hline

			               SemiContour                 & .73           & .75          & \textbf{.78}           \\
			               SemiContour-Seg        & \textbf{.74}  & \textbf{.77} & .76           \\ 
			             
		\end{tabularx}
	\end{center} \vspace{-0.6cm}
\end{table}

\begin{table}[t]
	\caption{Contour detection results of SemiContour on BSDS500 that is trained with different number of labeled training  images.}\label{table:bsds-variant}
	\vspace{-0.25cm}
	\begin{center}
		\begin{tabularx}{.38\textwidth}{c|YYY}
		{\small \# of Labeled Images}        & ODS  & OIS  & AP   \\ \specialrule{1.5pt}{0pt}{0pt}
			3  & .728 & .747 & .776 \\
			10 & .732 & .753 & .782  \\
			20 & .734 & .755 & .784 \\
			50 & .736 & .758 & .787 \\ 
		\end{tabularx}
	\end{center} \vspace{-0.7cm}
\end{table}

\begin{figure}[t]
	\begin{center}
		\subfigure[BSDS500]{\includegraphics[width=0.49\linewidth]{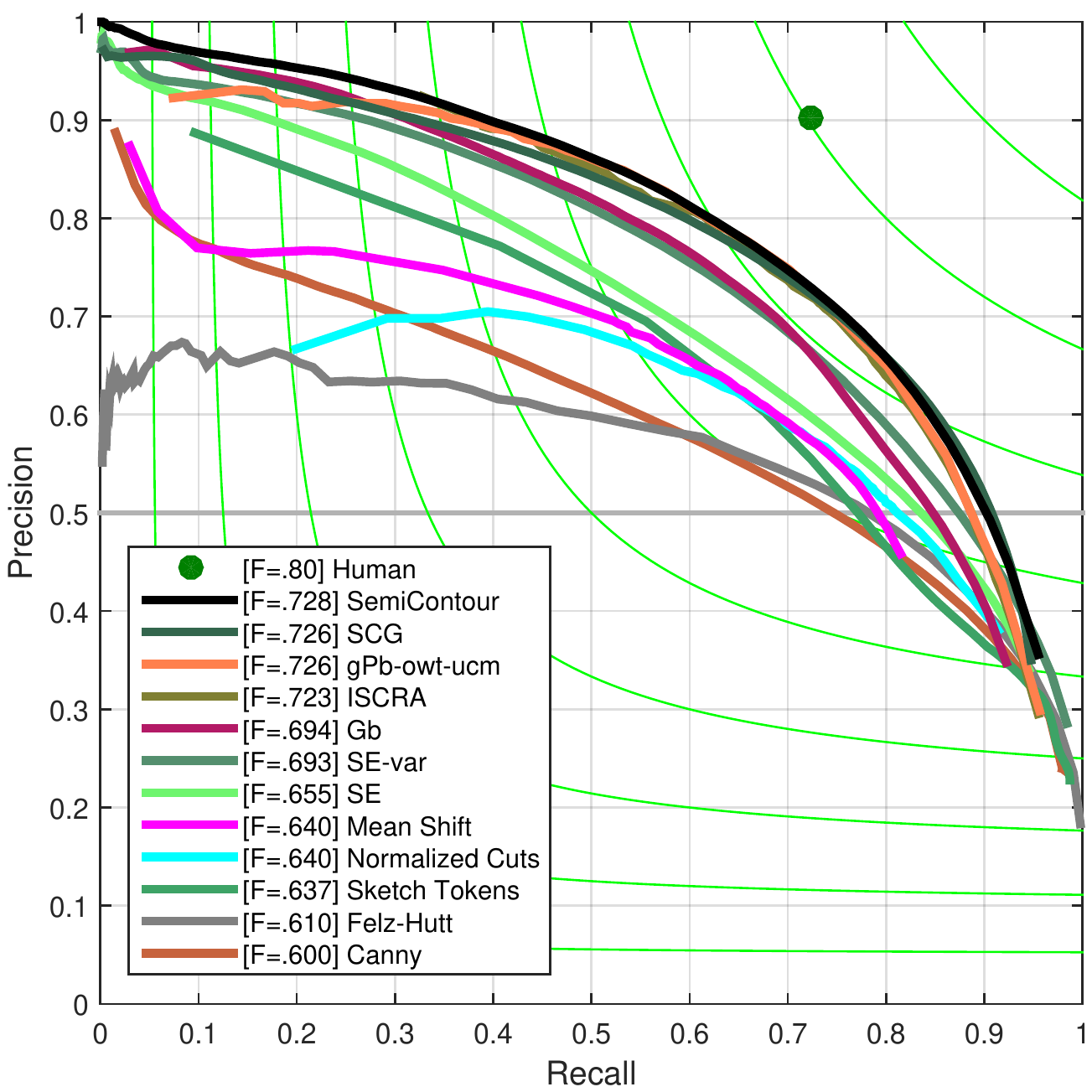}}
		\subfigure[NYUD]{\includegraphics[width=0.49\linewidth]{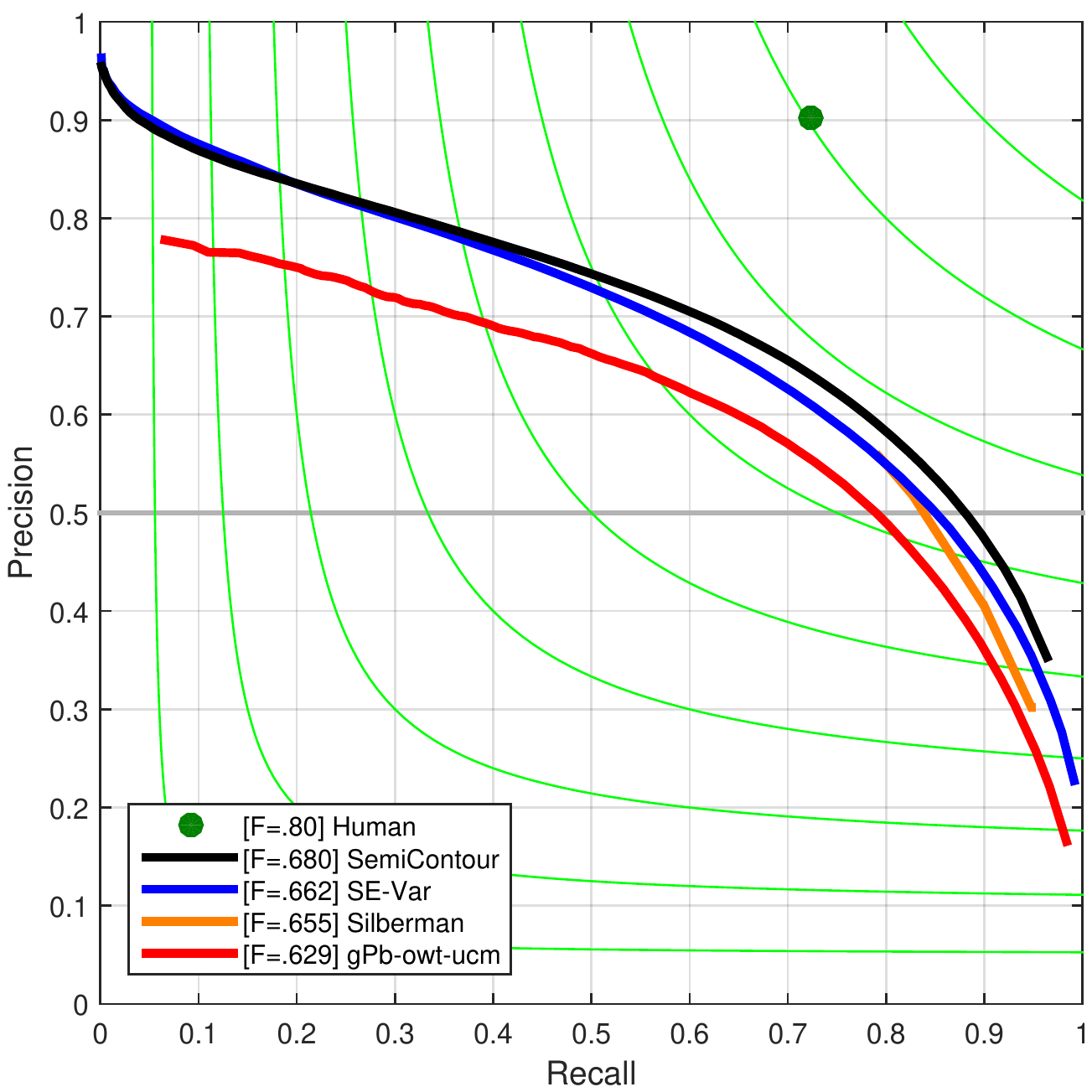}}
	\end{center}
	\vspace{-0.4cm}
	\caption{Precision/recall curves on BSDS500 and NYUD.} \vspace{-0.4cm}\label{fig:bsds}
\end{figure}

\vspace{0.1em}
\noindent
\textbf{BSDS500:} BSDS500 \cite{arbelaez2011contour} has been widely used as a benchmark for contour detection methods, including $200$ training, $100$ validation, and $200$ test images.  Our method uses $3$ labeled training images in the labeled set; the rest $197$ images are included in the unlabeled set. Table \ref{table:bsds} and Figure \ref{fig:bsds}(a) compare our method with several other methods\footnote{We carefully check every step when re-training their model and keep the other parameters default.}. 

In order to compare with supervised methods, we provide the performance with $3$ as well as with all $200$ labeled training images (comparative results with 200 images are obtained from the authors' original papers). As we can see, the proposed SemiContour method produces similar results as supervised methods using 200 training images, but outperforms all the unsupervised methods and supervised methods with $3$ labeled training images. The performance of all supervised approaches except SCG significantly decreases with only $3$ labeled training images. Specifically, compared with the SE-Var (an improved version of SE), our method exhibits $4$-point higher ODS and $4$-point higher AP. The gPb-owt-ucm and SCG, which merely replaces the local contrast estimation of the former that does not rely on many labeled images, exhibit close performance to ours, but our PR curve still shows higher precision with the same recall rates. In terms of efficiency, our method is hundreds of times faster than these two. For a $420\times 320$ image, SemiContour runs within $0.89$s, while gPb-owt-ucm and SCG require $240$s and $280$s, respectively. Several qualitative example results are shown in Figure \ref{fig:edge_vis}. In addition, we also show the experimental results of the proposed method using a different number of labeled images in Table \ref{table:bsds-variant}. 

\begin{figure}[t]
	\begin{center}
		\subfigure{\includegraphics[width=0.32\linewidth]{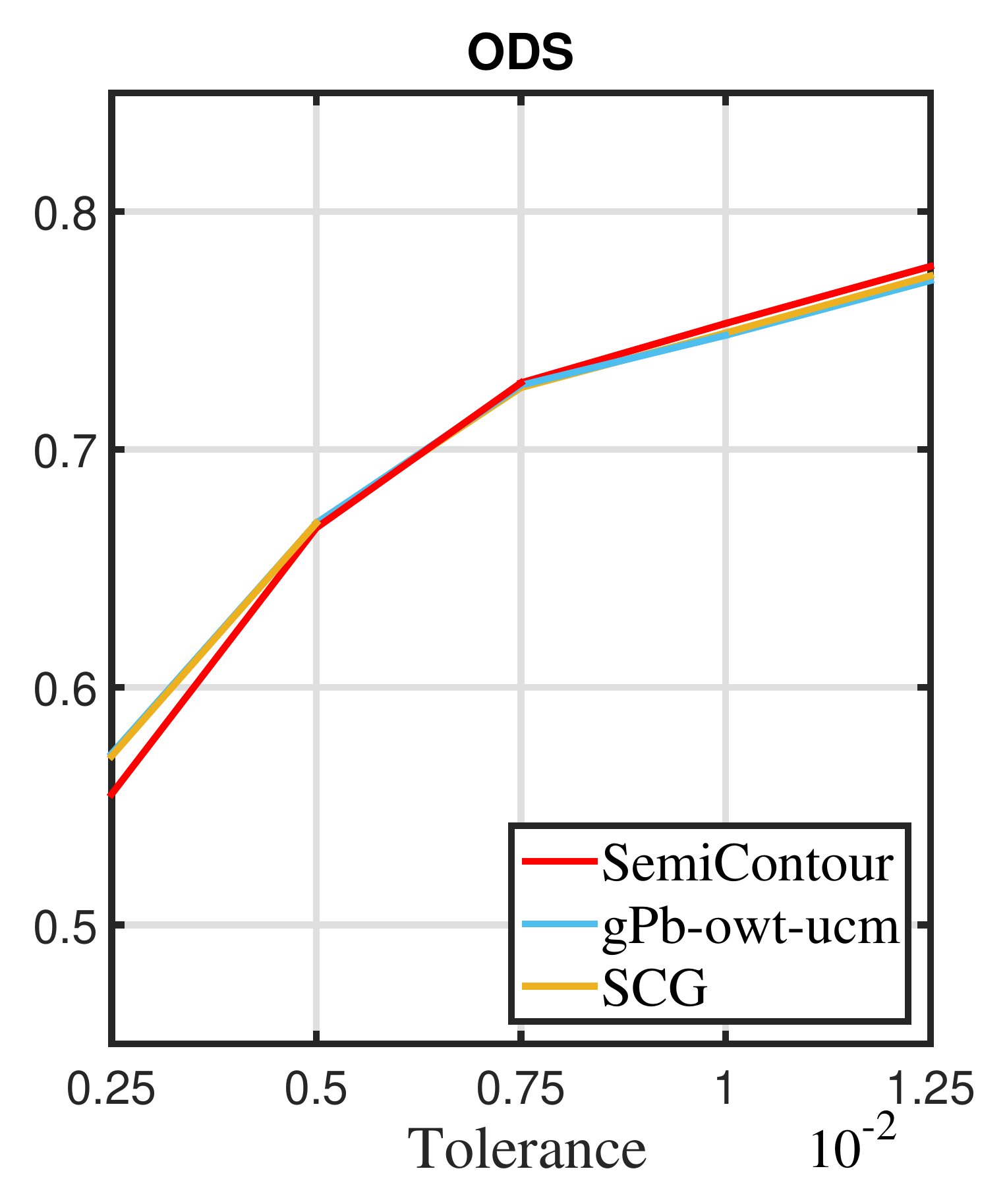}}
		\subfigure{\includegraphics[width=0.32\linewidth]{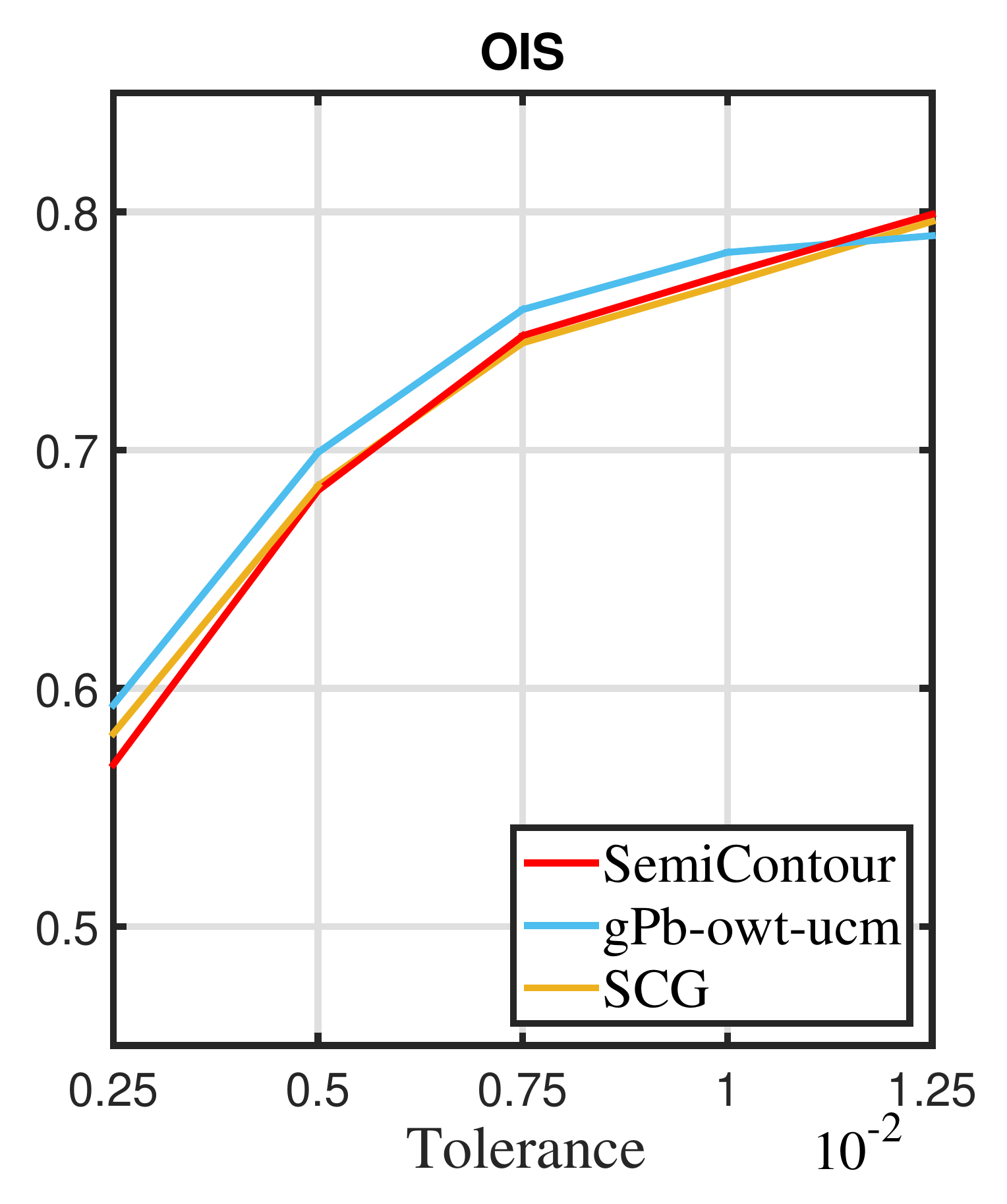}}
		\subfigure{\includegraphics[width=0.32\linewidth]{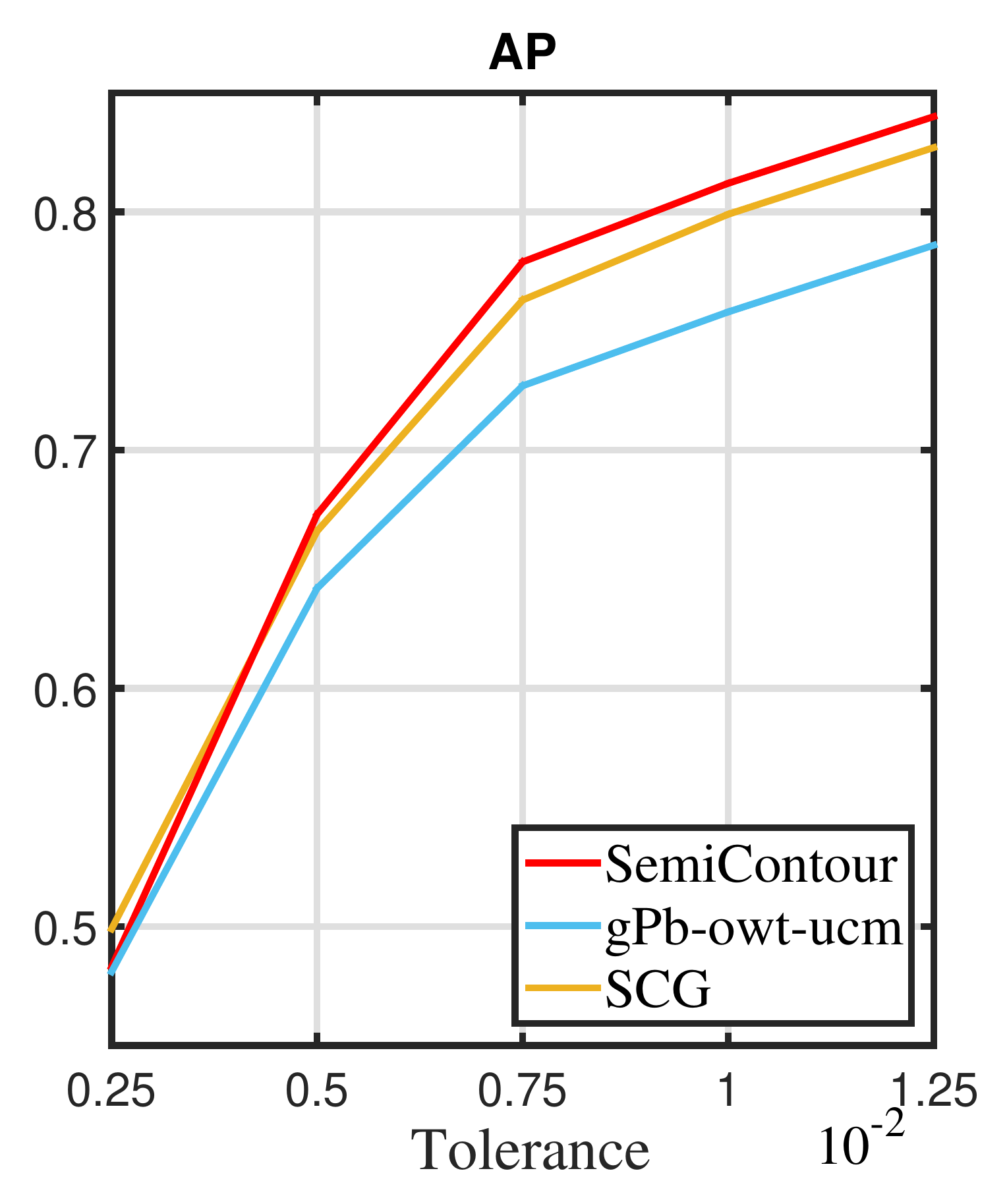}}
	\end{center}
	\vspace{-0.4cm}
	\caption{The comparative performance by varying tolerance thresholds (maximized pixel distance allowed when matching the estimated contours to ground-truth). SemiContour slightly underperforms SCG and gPb-owt-ucm at stringent thresholds due to some skewed localizations. However, it outperforms both, especially in AP measurement, with the slack thresholds, which means that SemiContour is less likely to miss real contours. } \vspace{-0.6cm} \label{fig:varthres}
\end{figure}

We find that the estimated structured labels of \textit{u-tokens} sometimes might cause skewed localization at exact contour position. However, our method is less likely to miss real contours, as shown in Figure \ref{fig:varthres}. Precise contour localization is necessary but less important in applications such as object detection and scene understanding.

\begin{table}[b]\vspace{-.5cm}
	\caption{Segmentation results on BSDS500. Evaluation criteria is described in \cite{arbelaez2011contour}. Note that we only use three labeled image to train the proposed SemiContour method.}\label{table:bsds-seg}
	\vspace{-0.3cm}
	\begin{center}
		\begin{tabularx}{.40\textwidth}{c|YY|YY}
			                                                                                                         & \multicolumn{2}{c|}{{\small Cover}} & \multicolumn{2}{c}{{\small PRI}} \\ \cline{2-5}
			                                                   ~                                                     & {\small ODS} & {\small OIS}         & {\small ODS}  & {\small OIS}     \\
			\specialrule{1.5pt}{0pt}{0pt}
			red-spectral \cite{taylor2013towards} & .56          & .62                  & .81           & .85              \\
			                                  gPb-owt-ucm \cite{arbelaez2011contour}                                  & .59          & \textbf{.65}                  & .83           & \textbf{.86}              \\
			                                     DC \cite{donoser2014discrete}                                       & .58          & .63                  & .82           & .85              \\ \hline
			                                            SemiContour-Seg                                              & \textbf{.59} & .64         & \textbf{.83 } & .85
		\end{tabularx}
	\end{center} \vspace{-0.5cm}
\end{table}

\vspace{0.2em}

We also test the performance of using the proposed SemiCoutour method for segmentation. After contour detections using SemiContour, multiscale-UCM \cite{arbelaez2014multiscale} is applied onto the generated contour images to generate the segmentation results (denoted as SemiContour-Seg in our experiments). We compare SemiContour-Seg with several state-of-the-art methods. The results are shown in Figure \ref{fig:edge_vis} and Table \ref{table:bsds-seg}. SemiContour-Seg also improves the contour detection performance as shown in Table \ref{table:bsds}.
\vspace{0.3em}

\noindent
\textbf{NYUD:}
NYUD contains 1449 RGB-D images. We follow \cite{dollar2013structured} to perform the experiment setup. The dataset is splited into 381 training, 414 validation, and 654 testing images. To conduct RGB-D contour detection, we treat the depth image as an extra feature channel, and thus the dictionary basis has five channels, and the feature channels for SRF are increased by 11 \cite{dollar2013structured}.  
We use $10$ images in the labeled set with the rest $371$ images in the unlabeled set. The comparison results are shown in Table \ref{table:nyud} and Figure \ref{fig:bsds}(b). We can observe that SemiContour with only 10 training images produces superior results than supervised methods trained with 10 images, and also provides competitive results with supervised methods trained using all 381 labeled data.
\vspace{0.3em}

\begin{table}[t]
	\caption{Contour detection results on NYUD. In the first column, from top to bottom, the first block is unsupervised method, the second block is supervised methods, and the third block is our method. For supervised methods (second block), we show the performance using both $10$ and $381$ training images (shown in $(\cdot)$). } \label{table:nyud}
	\vspace{-0.5em}
	\begin{center}
		\begin{tabularx}{.40\textwidth}{c|YYY}
			~                     & ODS          & OIS          & AP           \\ \specialrule{1.5pt}{0pt}{0pt}
			gPb-owt-ucm \cite{arbelaez2011contour}   & .63          & .66          & .56          \\ \hline
			Siberman \cite{silberman2012indoor}              & - (.65)       & - (.66)       & - (.29)       \\

			SE-Var \cite{dollar2013structured}     & .66(.69)     & .68(.71)     & .68(.72)     \\ \hline
			SemiContour                & \textbf{.68} & \textbf{.70} & \textbf{.69} \\ 
		\end{tabularx}
	\end{center}\vspace{-0.6cm}
	
\end{table}

\begin{table}[t]
	\caption{Cross-dataset generalization results. The first column indicates the training/testing dataset settings that we used. SemiContour outperforms SE-Var on both settings.}\label{table:cross}
	\vspace{-0.3cm}
	\begin{center}
		\begin{tabularx}{.45\textwidth}{c|c|ccc}
			\small                              &             &     ODS      &      OIS      &      AP       \\
			\specialrule{1.5pt}{0pt}{0pt}
			
			\multirow{2}{*}{NYUD/BSDS}										 &   SE-Var    &     .73      &      .74      &      .77      \\
			& SemiContour & \textbf{.73} & \textbf{.75} & \textbf{.78} \\ \hline
			\multirow{2}{*}{BSDS/NYUD}                    &   SE-Var    &     .64      &      .66      &      .63      \\
			& SemiContour & \textbf{.65} & \textbf{.66}  & \textbf{.63} \\ 
			%
			%
		\end{tabularx}
	\end{center} \vspace{-0.7cm}
\end{table}
\vspace{-.2cm}
\subsection{Cross-dataset generalization results}
One advantage of the proposed SemiContour is that it can improve the generalization ability of contour detection by incorporating unlabeled data from the target dataset domain. To validate this, we perform a cross-dataset experiment on BSDS500 and NYUD. The two datasets exhibit significant visual variations. NYUD contains various indoor scenes under different lighting conditions, and BSDS500 contains outdoor scenes. We use one dataset as the labeled set and another as the unlabeled set. The rest experiment setup is the same as SE-Var \cite{dollar2015pami}. We compare SemiContour with SE-Var in Table \ref{table:cross}\footnote{Later on, we conducted an extra experiment to augment 200 labeled training images of BSDS with 100 unlabeled images of NYUD to improve the testing results of BSDS. Our method achieves (.752ODS, .786OIS, .792AP), compared with SE-Var's results (.743ODS, .763OIS, .788AP), both with totally 1 million training tokens.}.

These experiments validate the strong generalization ability and the robustness of the proposed SemiContour method, which indicates a strong noise resistance of the method even when we incorporate \textit{u-tokens} from a different image domain.

\vspace{-.1cm}
\subsection{Efficiency of the proposed fast sparse coding}
\vspace{-.1cm}
\label{sec:sc}
The running time of our novel sparse coding algorithm is determined by the steps of basis selection and reconstruction error minimization. The former step needs $O(d\scdot V)$ to compute $V$ scores and $O(V\scdot K)$ to select the $K$ bases, and the latter reconstruction error minimization step needs $O(d\scdot K^2)$ with a $d\times K$ dictionary. Therefore, the total time complexity is $\max\left(O(d\scdot V),O(d\scdot K^2)\right)$, usually $O(d\scdot V)$ because $K$ is much smaller than $V$ in practice.

\begin{figure}[t]
	\begin{center}
		\includegraphics[width=0.7\linewidth]{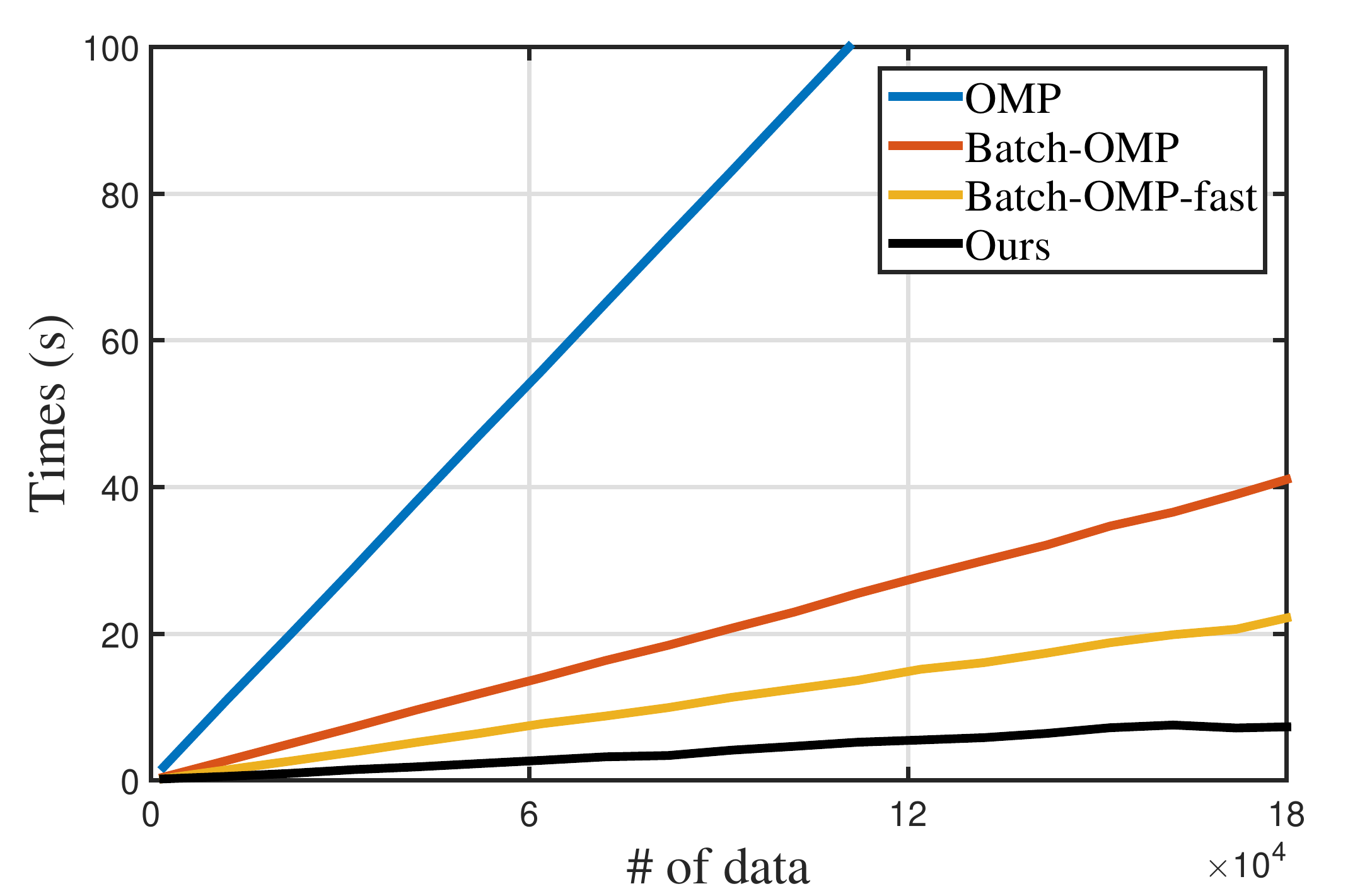}
	\end{center}
	\vspace{-0.9em}
	\caption{Runtime comparison results. The dictionary size is $576\times512$ and the sparsity $K{=}6$. Our method significantly outperforms the others as the number of target data increases.} \vspace{-0.6cm} \label{fig:timecomp}
\end{figure}

We compare our fast sparse coding solver with several algorithms in Figure \ref{fig:timecomp}. Most of existing sparse coding algorithms suffer from computational expensive iterations. We only choose several popular ones to compare with our algorithm, including OMP \cite{pati1993orthogonal}, Batch-OMP \cite{pati1993orthogonal} and its faster version (Batch-OMP-fast). All of these comparative algorithms contain highly optimized implementations and our algorithm is a simple Matlab implementation. We observe that our fast sparse coding algorithm obtains the same results as the others in terms of the final contour detection accuracy, but it is significantly faster than the others. Since the computation of each target data is independent, an additional benefit is that the proposed algorithm can be easily parallelized. All algorithms are tested on an Intel i7@3.60GHz$\times$6 cores and 32GB RAM machine.

\vspace{-.1cm}
\section{Conclusions}
\vspace{-.1cm}
In this paper, we present a novel semi-supervised structured ensemble learning method for contour detection. Specifically, our approach trains an effective contour detector based on structured random forests (SRF). We take advantage of unlabeled data to conduct better node splitting of SRF using sparse representation techniques, whose procedures are embedded in the overall SRF training. In order to increase the scalability of sparse coding to extensive target data, we have proposed a fast and robust sparse coding algorithm. Compared with many existing literatures, our method provides superior testing results.

{\small
\bibliographystyle{ieee}
\bibliography{references}

\begin{thebibliography}{10}\itemsep=-1pt

\bibitem{aharon2006img}
M.~Aharon, M.~Elad, and A.~Bruckstein.
\newblock K-svd: An algorithm for designing overcomplete dictionaries for
  sparse representation.
\newblock {\em IEEE Trans. on Signal Processing}, 54(11):4311--4322, 2006.

\bibitem{arbelaez2011contour}
P.~Arbelaez, M.~Maire, C.~Fowlkes, and J.~Malik.
\newblock Contour detection and hierarchical image segmentation.
\newblock {\em PAMI}, 33(5):898--916, 2011.

\bibitem{arbelaez2014multiscale}
P.~Arbelaez, J.~Pont-Tuset, J.~Barron, F.~Marques, and J.~Malik.
\newblock Multiscale combinatorial grouping.
\newblock In {\em CVPR}, pages 328--335, 2014.

\bibitem{bertasius2014deepedge}
G.~Bertasius, J.~Shi, and L.~Torresani.
\newblock Deepedge: A multi-scale bifurcated deep network for top-down contour
  detection.
\newblock {\em arXiv preprint arXiv:1412.1123}, 2014.

\bibitem{bo2013multipath}
L.~Bo, X.~Ren, and D.~Fox.
\newblock Multipath sparse coding using hierarchical matching pursuit.
\newblock In {\em CVPR}, pages 660--667, 2013.

\bibitem{breiman2001random}
L.~Breiman.
\newblock Random forests.
\newblock {\em Machine learning}, 45:5--32, 2001.

\bibitem{canny1986computational}
J.~Canny.
\newblock A computational approach to edge detection.
\newblock {\em PAMI}, (6):679--698, 1986.

\bibitem{chapelle2006semi}
O.~Chapelle, B.~Sch{\"o}lkopf, A.~Zien, et~al.
\newblock Semi-supervised learning.
\newblock 2006.

\bibitem{comaniciu2002mean}
D.~Comaniciu and P.~Meer.
\newblock Mean shift: A robust approach toward feature space analysis.
\newblock {\em PAMI}, 24(5):603--619, 2002.

\bibitem{cour2005spectral}
T.~Cour, F.~Benezit, and J.~Shi.
\newblock Spectral segmentation with multiscale graph decomposition.
\newblock In {\em CVPR}, volume~2, pages 1124--1131, 2005.

\bibitem{criminisi2012decision}
A.~Criminisi, J.~Shotton, and E.~Konukoglu.
\newblock {\em Decision forests: A unified framework for classification,
  regression, density estimation, manifold learning and semi-supervised
  learning}, volume~7.
\newblock 2012.

\bibitem{dollar2015pami}
P.~Doll{\'a}r and C.~Zitnick.
\newblock Fast edge detection using structured forests.
\newblock {\em PAMI}, 2015.

\bibitem{dollar2013structured}
P.~Doll{\'a}r and C.~L. Zitnick.
\newblock Structured forests for fast edge detection.
\newblock In {\em ICCV}, pages 1841--1848, 2013.

\bibitem{donoser2014discrete}
M.~Donoser and D.~Schmalstieg.
\newblock Discrete-continuous gradient orientation estimation for faster image
  segmentation.
\newblock In {\em CVPR}, pages 3158--3165, 2014.

\bibitem{felzenszwalb2004efficient}
P.~F. Felzenszwalb and D.~P. Huttenlocher.
\newblock Efficient graph-based image segmentation.
\newblock {\em IJCV}, 59(2):167--181, 2004.

\bibitem{ganin2014n}
Y.~Ganin and V.~Lempitsky.
\newblock N\^{} 4-fields: Neural network nearest neighbor fields for image
  transforms.
\newblock In {\em ACCV}, pages 536--551. 2014.

\bibitem{kontschieder2011structured}
P.~Kontschieder, S.~R. Bulo, H.~Bischof, and M.~Pelillo.
\newblock Structured class-labels in random forests for semantic image
  labelling.
\newblock In {\em ICCV}, pages 2190--2197, 2011.

\bibitem{leistner2009semi}
C.~Leistner, A.~Saffari, J.~Santner, and H.~Bischof.
\newblock Semi-supervised random forests.
\newblock In {\em ICCV}, pages 506--513, 2009.

\bibitem{leordeanu2014generalized}
M.~Leordeanu, R.~Sukthankar, and C.~Sminchisescu.
\newblock Generalized boundaries from multiple image interpretations.
\newblock {\em PAMI}, 36(7):1312--1324, 2014.

\bibitem{lim2013sketch}
J.~J. Lim, C.~L. Zitnick, and P.~Doll{\'a}r.
\newblock Sketch tokens: A learned mid-level representation for contour and
  object detection.
\newblock In {\em CVPR}, pages 3158--3165, 2013.

\bibitem{liu2011robust}
B.~Liu, J.~Huang, L.~Yang, and C.~Kulikowsk.
\newblock Robust tracking using local sparse appearance model and k-selection.
\newblock In {\em CVPR}, pages 1313--1320, 2011.

\bibitem{fujun2015maccai}
F.~Liu, F.~Xing, Z.~Zhang, M.~Mcgough, and L.~Yang.
\newblock Robust muscle cell quantification using structured edge detection and
  hierarchical segmentation.
\newblock In {\em MICCAI}, pages 324--331, 2015.

\bibitem{liu2013semi}
X.~Liu, M.~Song, D.~Tao, Z.~Liu, L.~Zhang, C.~Chen, and J.~Bu.
\newblock Semi-supervised node splitting for random forest construction.
\newblock In {\em CVPR}, pages 492--499, 2013.

\bibitem{maire2014reconstructive}
M.~Maire, X.~Y. Stella, and P.~Perona.
\newblock Reconstructive sparse code transfer for contour detection and
  semantic labeling.
\newblock In {\em ACCV}, pages 273--287. 2014.

\bibitem{myers2015affordance}
A.~Myers, C.~L. Teo, C.~Ferm{\"u}ller, and Y.~Aloimonos.
\newblock Affordance detection of tool parts from geometric features.
\newblock In {\em ICRA}, 2015.

\bibitem{pati1993orthogonal}
Y.~C. Pati, R.~Rezaiifar, and P.~Krishnaprasad.
\newblock Orthogonal matching pursuit: Recursive function approximation with
  applications to wavelet decomposition.
\newblock In {\em Asilomar Conference on Signals, Systems and Computers}, pages
  40--44, 1993.

\bibitem{ren2006figure}
X.~Ren, C.~C. Fowlkes, and J.~Malik.
\newblock Figure/ground assignment in natural images.
\newblock In {\em ECCV}, pages 614--627. 2006.

\bibitem{ren2013image}
Z.~Ren and G.~Shakhnarovich.
\newblock Image segmentation by cascaded region agglomeration.
\newblock In {\em CVPR}, pages 2011--2018, 2013.

\bibitem{shen2015deepcontour}
W.~Shen, X.~Wang, Y.~Wang, X.~Bai, and Z.~Zhang.
\newblock Deepcontour: A deep convolutional feature learned by positive-sharing
  loss for contour detection.
\newblock In {\em CVPR}, pages 3982--3991, 2015.

\bibitem{silberman2012indoor}
N.~Silberman, D.~Hoiem, P.~Kohli, and R.~Fergus.
\newblock Indoor segmentation and support inference from rgbd images.
\newblock In {\em ECCV}, pages 746--760. 2012.

\bibitem{tang2007co}
F.~Tang, S.~Brennan, Q.~Zhao, and H.~Tao.
\newblock Co-tracking using semi-supervised support vector machines.
\newblock In {\em ICCV}, pages 1--8, 2007.

\bibitem{taylor2013towards}
C.~J. Taylor.
\newblock Towards fast and accurate segmentation.
\newblock In {\em CVPR}, pages 1916--1922, 2013.

\bibitem{teo2015fast}
C.~L. Teo, C.~Ferm{\"u}ller, and Y.~Aloimonos.
\newblock Fast 2d border ownership assignment.
\newblock In {\em CVPR}, pages 5117--5125, 2015.

\bibitem{wang2010locality}
J.~Wang, J.~Yang, K.~Yu, F.~Lv, T.~Huang, and Y.~Gong.
\newblock Locality-constrained linear coding for image classification.
\newblock In {\em CVPR}, pages 3360--3367, 2010.

\bibitem{xiaofeng2012discriminatively}
R.~Xiaofeng and L.~Bo.
\newblock Discriminatively trained sparse code gradients for contour detection.
\newblock In {\em Advances in NIPS}, pages 584--592, 2012.

\bibitem{xie2015holistically}
S.~Xie and Z.~Tu.
\newblock Holistically-nested edge detection.
\newblock {\em arXiv preprint arXiv:1504.06375}, 2015.

\bibitem{yang2012beyond}
J.~Yang, L.~Zhang, Y.~Xu, and J.-y. Yang.
\newblock Beyond sparsity: The role of l 1-optimizer in pattern classification.
\newblock {\em Pattern Recognition}, 45(3):1104--1118, 2012.

\end{thebibliography}
}

\end{document}